\documentclass{article}

\usepackage{microtype}
\usepackage{graphicx}
\usepackage{subfigure}
\usepackage{booktabs} 
\usepackage{makecell} 

\usepackage{hyperref}

\usepackage[arxiv]{icml2025}

\usepackage{amsmath}
\usepackage{amssymb}
\usepackage{mathtools}
\usepackage{amsthm}

\usepackage{float}

\definecolor{Green}{HTML}{3d9f3c}

\usepackage[capitalize,noabbrev]{cleveref}

\theoremstyle{plain}

\theoremstyle{definition}

\theoremstyle{remark}

\usepackage[textsize=tiny]{todonotes}

\icmltitlerunning{Describe Anything in Medical Images}

\begin{document}

\twocolumn[
\icmltitle{Describe Anything in Medical Images}



\icmlsetsymbol{equal}{\dag}
\icmlsetsymbol{corresponding}{*}

\begin{icmlauthorlist}
\icmlauthor{Xi Xiao}{equal,1}
\icmlauthor{Yunbei Zhang}{equal,2}
\icmlauthor{Thanh-Huy Nguyen}{equal,3}
\icmlauthor{Ba-Thinh Lam}{6}
\icmlauthor{Janet Wang}{2}
\icmlauthor{Lin Zhao}{5}

\icmlauthor{Jihun Hamm}{2}
\icmlauthor{Tianyang Wang}{1}
\icmlauthor{Xingjian Li}{3}
\icmlauthor{Xiao Wang}{4}
\icmlauthor{Hao Xu}{7}
\icmlauthor{Tianming Liu}{corresponding,8}
\icmlauthor{Min Xu}{corresponding,3}
\end{icmlauthorlist}

\icmlaffiliation{1}{University of Alabama at Birmingham}
\icmlaffiliation{2}{Tulane University}
\icmlaffiliation{3}{Carnegie Mellon University}
\icmlaffiliation{4}{Oak Ridge National Laboratory}
\icmlaffiliation{5}{Northeastern University}
\icmlaffiliation{6}{AI VIETNAM}
\icmlaffiliation{7}{Harvard Medical School}
\icmlaffiliation{8}{University of Georgia}

\icmlcorrespondingauthor{Tianming Liu}{tliu@cs.uga.edu}
\icmlcorrespondingauthor{Min Xu}{mxu1@cs.cmu.edu}

\icmlkeywords{Machine Learning, ICML}

\vskip 0.3in
]



\printAffiliationsAndNotice{\icmlEqualContribution} 

\begin{abstract}
Localized image captioning has made significant progress with models like the Describe Anything Model (DAM), which can generate detailed region-specific descriptions without explicit region-text supervision. 
However, such capabilities have yet to be widely applied to specialized domains like medical imaging, where diagnostic interpretation relies on subtle regional findings rather than global understanding. 
To mitigate this gap, we propose \textbf{MedDAM}, the first comprehensive framework leveraging large vision-language models for \textit{region-specific captioning in medical images}. MedDAM employs medical expert-designed prompts tailored to specific imaging modalities and establishes a robust evaluation benchmark comprising a customized assessment protocol, data pre-processing pipeline, and specialized QA template library. This benchmark evaluates both MedDAM and other adaptable large vision-language models, focusing on clinical factuality through attribute-level verification tasks—elegantly circumventing the absence of ground-truth region-caption pairs in medical datasets. Extensive experiments on the VinDr-CXR, LIDC-IDRI, and SkinCon datasets demonstrate MedDAM's superiority over leading peers (including GPT-4o, Claude 3.7 Sonnet, LLaMA-3.2 Vision, Qwen2.5-VL, GPT-4Rol, and OMG-LLaVA) in the task, revealing the importance of region-level semantic alignment in medical image understanding and establishing MedDAM as a promising foundation for clinical vision-language integration.
\end{abstract}

\section{Introduction}

Vision-language models (VLMs) have made remarkable strides in generating natural language descriptions of visual content. Traditional captioning models often focus on interpreting entire images~\cite{vinyals2015show,anderson2018bottom}, but many real-world applications require a deep understanding and precise descriptions of fine-grained and localized regions. Recent advances in region-level captioning~\cite{johnson2016densecap,li2022blip,li2023blip,lian2025anythingdetailedlocalizedimage} have introduced models capable of producing detailed region-specific descriptions without relying on explicit region-text supervision. Among them, the Describe Anything Model (DAM)~\cite{lian2025anythingdetailedlocalizedimage} is particularly notable, leveraging focal prompting, a localized visual backbone, and a self-supervised data pipeline (DLC-SDP) to achieve state-of-the-art performance on localized captioning tasks for natural images.

Though promising for natural images, the extension of such models to medical images remains unexplored. In clinical practice, diagnostic interpretation could highly rely on subtle localized findings rather than a holistic understanding. Detailed descriptions of localized regions—capturing crucial features such as location, morphology, density, and boundary characteristics—are essential for accurate diagnosis and effective communication among healthcare professionals. However, existing medical image captioning methods~\cite{jing2018automatic,chen2020generating,wang2022medclip,huang2023radclip} are developed to generate image-level descriptions, failing to deliver  fine-grained region-specific  descriptions urgently needed by clinicians (e.g., radiologists) for diagnosis. 

Unlike segmentation models such as MedSAM~\cite{ma2023medsam} and MedSAM2~\cite{ma2024medsam2} that predict explicit region masks, the  recent breakthrough DAM \cite{lian2025anythingdetailedlocalizedimage} generates free-form textual descriptions without predefined classes or structured labels~\cite{wang2024multi,xiao2024hgtdp}. However, applying DAM directly to medical images introduces unique challenges. Moreover, medical datasets rarely provide region-specific captions, rendering conventional captioning evaluation metrics (e.g., BLEU \cite{papineni2002bleu}, ROUGE \cite{lin2004rouge}) inapplicable. This raises a critical question: \emph{Can a general localized captioning model like DAM generalize to medical images, and how should its outputs be evaluated in the absence of ground-truth region-specific descriptions?}

To answer it, we introduce our \textbf{MedDAM} framework, a practical solution for region-specific captioning across diverse medical images. It integrates three key components: (1) medical expert-designed prompts tailored for different organs and imaging modalities (chest X-ray, lung CT scan, dermatology, etc.), enabling appropriate domain-specific query; (2) a flexible region-of-interest detection pipeline that leverages existing segmentation models when boundary boxes or masks are unavailable in the original dataset, making it adaptable to any medical imaging collection; and (3) a specialized evaluation benchmark that assesses the quality and accuracy of detailed localized captioning without requiring reference captions through attribute-level verification tasks. This unified framework performs effectively across any available medical imaging datasets regardless of domain or availability of pre-existing segmentation annotations or captions, offering a versatile and robust framework for advancing region-specific understanding in medical images.
We conduct extensive evaluations on three clinically significant datasets: VinDr-CXR for chest radiography, LIDC-IDRI for lung CT imaging, and SkinCon for skin imaging, covering critical diagnostic tasks across different imaging modalities. Our main contributions are summarized as follows:
\begin{itemize}
    \item 
    We introduce \textbf{MedDAM}, a framework that enables region-specific captioning for medical images through expert-designed prompts and a specialized evaluation protocol for clinical applications.
    \item 
    We establish the first benchmark comparing MedDAM against leading large VLMs across chest radiography, lung CT, and dermatology, demonstrating its significant advantages in region-specific clinical understanding.

\end{itemize}

\section{Related Works}

\subsection{Localized Image Captioning}

Image captioning traditionally focuses on generating a global description that summarizes the salient content of an entire image~\cite{vinyals2015show,anderson2018bottom}. However, many applications demand fine-grained understanding of localized regions rather than holistic summaries. Early works such as DenseCap~\cite{johnson2016densecap} pioneered dense captioning, which jointly detects regions and generates region-level descriptions. Despite its impact, DenseCap relied heavily on supervised region proposals and densely annotated datasets, limiting its applicability in real-world scenarios. Recent advances in vision-language pretraining have improved the ability to align local visual features with textual representations. VinVL~\cite{zhang2021vinvl} boosted region-level captioning performance by strengthening object detectors and leveraging larger and higher-quality training corpora. BLIP~\cite{li2022blip} and BLIP-2~\cite{li2023blip} further advanced this field by using bootstrapped captions and lightweight bridging modules to enhance region-conditioned text generation without full supervision. Grounding DINO ~\cite{liu2023groundingdino} proposed end-to-end regional grounding by tightly coupling language prompts with visual object detection, enabling open-vocabulary and open-region localization. Nevertheless, these methods often rely on synthetic captions or global image-text pairs, significantly limiting their applications in medical image scenarios. A very recent advance, namely Describe Anything Model (DAM)~\cite{lian2025anythingdetailedlocalizedimage}, addresses this gap by introducing focal prompting and a localized backbone design, along with a self-supervised data pipeline (DLC-SDP) that automatically constructs detailed region-level pseudo-captions. DAM achieves state-of-the-art performance on localized captioning tasks without requiring region-text annotations, setting a new standard for detailed and scalable region understanding.

Despite promising, localized captioning has been predominantly explored in natural images. Extending such models to specialized domains, such as medical image analysis, remains challenging. In this work, we take the first step toward adapting localized captioning models (i.e., originally developed for natural images) to medical images.

\subsection{Medical Image Captioning}

Medical image captioning aims to automatically generate (e.g., radiology) reports or textual descriptions from medical images. 
Earlier approaches~\cite{jing2018automatic,chen2020generating} framed this task as an image-to-sequence problem, applying encoder-decoder architectures. 
Models such as R2Gen~\cite{chen2020generating} introduced memory-driven decoding strategies, while others incorporated hierarchical structures to better reflect the nature of the generated reports. With the availability of large-scale datasets such as CheXpert~\cite{irvin2019chexpert} and MIMIC-CXR~\cite{johnson2019mimic}, supervised report generation has become the major paradigm. More recently, contrastive vision-language pretraining has been adopted to align medical images with their associated reports. MedCLIP~\cite{wang2022medclip} adapts the CLIP framework \cite{radford2021learning} to report generation through knowledge-driven semantic objectives, while RadCLIP~\cite{huang2023radclip} introduces volumetric alignment techniques for 2D and 3D medical images. In parallel, segmentation foundation models such as MedSAM~\cite{ma2023medsam} and MedSAM2~\cite{ma2024medsam2} have enabled universal promptable segmentation across diverse imaging modalities, focusing on delineating anatomical structures and lesions, however, these models generate masks rather than textual descriptions, leaving the semantic interpretation of regions to human users.

\begin{figure}[t]
\centering
\includegraphics[width=\linewidth]{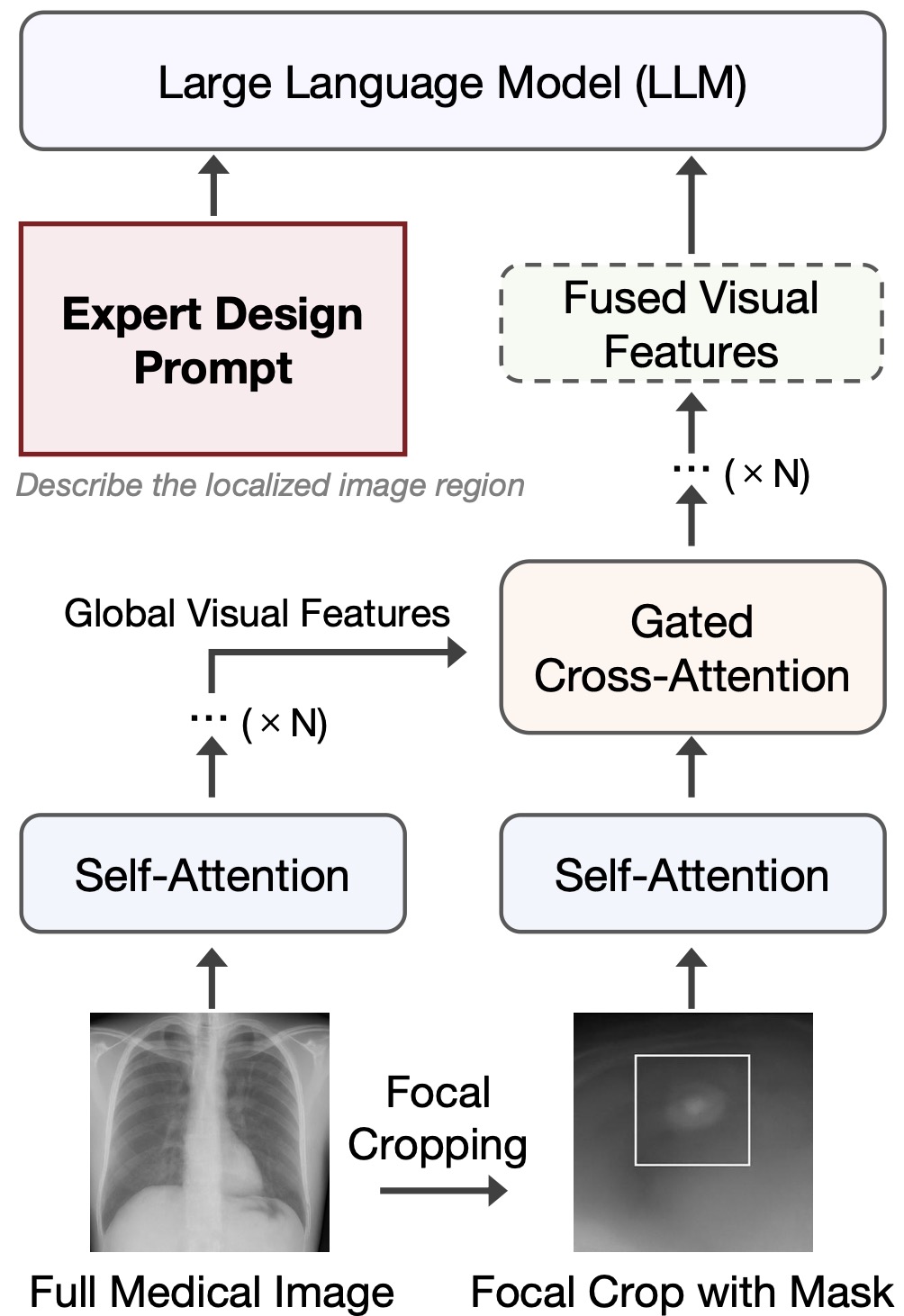} 
\caption{\textbf{Architecture of MedDAM.} MedDAM extends the recent breakthrough, i.e., Describe Anything framework \cite{lian2025anythingdetailedlocalizedimage}, to medical image understanding. A clinically focused region and its binary mask are used to generate a focal crop, which is embedded along with the full image, fusing global and regional features via gated cross-attention, while structured prompt tokens encode clinical objectives. Then, the resulting features are fed into a LLM to generate region-specific captions.}
\label{fig:medicaldam_architecture}
\end{figure}

While report generation models capture global image-level information, and segmentation models extract structural regions, the intermediate task of producing localized and detailed textual descriptions of specific regions remains underexplored. Recent efforts such as UniVLG~\cite{jain2025univlg} have unified multiple vision-language tasks, but still operate mainly at image level. To date, there exists no systematic study of localized region-level captioning for medical images. \textit{Our work fills this gap by realizing a deployable framework MedDAM (Fig. \ref{fig:medicaldam_architecture}) to facilitate the zero-shot transfer of natural-image-based localized captioning models, including the very recent breakthrough, namely DAM \cite{lian2025anythingdetailedlocalizedimage}, to medical image understanding.} 


\section{Framework and Evaluation}

\subsection{Datasets}

We evaluate the MedDAM framework (Fig. \ref{fig:medicaldam_architecture}) across three public medical image datasets spanning different imaging modalities and diagnostic contexts. These include chest radiographs, lung CT scans, and dermatological photographs, enabling a comprehensive assessment of the model’s generalization 
on different imaging modalities and organs. The datasets used are summarized in Table~\ref{tab:datasets}.

\begin{table*}[t]
\centering
\caption{Summary of the datasets used for evaluating MedDAM.}
\small
\label{tab:datasets}
\begin{tabular}{p{2.8cm}p{2.2cm}p{4.2cm}p{4.4cm}p{1.5cm}}
\toprule
\textbf{Dataset} & \textbf{Modality} & \textbf{Region Annotation (Source \& Type)} & \textbf{Region Sampling Strategy \& Evaluation Focus} & \textbf{Sample Size} \\
\midrule
VinDr-CXR~\cite{nguyen2022vindr} & Chest X-ray (CXR) & Bounding boxes (thoracic abnormalities); 2D bounding boxes & Use all annotated boxes with 10\% margin expansion; evaluate lesion localization, opacity patterns, consolidation, and effusion findings & 18,000 images \\
\midrule
LIDC-IDRI~\cite{armato2011lidc} & Lung CT scan (3D) & Segmentation masks (nodules); 2D bounding boxes from selected slices & Select slice with highest radiologist agreement; extract tight bounding boxes from 2D mask slices; evaluate nodule morphology including margin, spiculation, and internal texture & 1,018 scans \\
\midrule
SkinCon~\cite{daneshjou2022skincon} & Clinical photography & Bounding boxes (cutaneous lesions); 2D bounding boxes & Use all annotated lesion regions with margin padding; evaluate dermatological attributes such as shape, color, border regularity, and surface texture & 3,000 images \\
\bottomrule
\end{tabular}
\end{table*}


VinDr-CXR~\cite{nguyen2022vindr} is a large-scale chest X-ray dataset including 18,000 frontal radiographs annotated with bounding boxes and labels for 14 common thoracic abnormalities, such as consolidation, lung opacity, and pleural effusion. It provides a valuable benchmark for evaluating localized captioning in 2D grayscale imaging, where visual abnormalities are often subtle and spatially diffuse.

LIDC-IDRI~\cite{armato2011lidc} consists of 1,018 lung CT scans annotated with detailed segmentation masks and semantic attributes for pulmonary nodules, including size, margin sharpness, density, and spiculation. This dataset introduces the challenge of generating localized descriptions within a 3D volumetric image, where fine-grained morphological characteristics play a key role in diagnostic interpretation.


SkinCon~\cite{daneshjou2022skincon} is a dermatology dataset densely annotated by expert dermatologists, consisting of 3,230 skin lesion images selected from the Fitzpatrick17k dataset, with up to 48 clinical concepts, and reflecting visual characteristics commonly used in clinical skin disease diagnosis.


\subsection{Region Sampling and Prompt Construction}
\label{sampling_construct}

Since MedDAM generates descriptions conditioned on localized visual regions, it is important to carefully construct region prompts that reflect clinically meaningful areas while maintaining consistency across datasets. Although the DAM model itself does not rely on manual annotations during inference, we utilize existing annotations (e.g., bounding boxes or segmentation masks), only for evaluation to identify diagnostically relevant regions for prompting. This ensures that the regions of interest used in benchmarking are both clinically grounded and comparable across models. We then standardize the extracted regions (e.g., margin padding, aspect ratio preservation) as input patches to DAM's focal prompting mechanism.

In VinDr-CXR, we directly utilize the provided bounding box annotations, each corresponding to a localized thoracic abnormality. To ensure sufficient contextual information while focusing on the target region, we enlarge each bounding box by a fixed margin of 10\%. LIDC-IDRI provides 3D segmentation masks for pulmonary nodules. We extract 2D slices containing nodules and generate bounding boxes from the segmentation masks. For each nodule, we select the axial slice with the most representative appearance (highest radiologist consensus) and crop the minimal enclosed region. For SkinCon, since the dataset doesn't contain any masks or bounding boxes, we implement a region-of-interest detection pipeline adapted from \cite{wang2024achieving} to identify dermatological lesions. Unlike \cite{wang2024achieving}, which uses ROI for precise lesion classification, our approach only requires approximate bounding boxes for region-specific captioning purposes. We then extend these detected regions with a slight margin to capture surrounding skin texture variations, providing sufficient context for generating detailed region-specific descriptions. This flexible ROI detection strategy demonstrates how MedDAM can be extended to any medical imaging dataset lacking ground-truth annotations, regardless of imaging modality or anatomical structure, by leveraging existing segmentation techniques to enable region-specific captioning even without manual annotations.


Across all the datasets, the extracted region prompts are processed through DAM's focal prompting mechanism, where a high-resolution crop of the region is combined with the full-scale image input. This ensures that the model focuses on the target area while preserving relevant global cues that may influence clinical interpretation.
Table~\ref{tab:region_prompting} summarizes the region sampling strategies and preprocessing steps across the three datasets.

\begin{table*}[t]
\centering
\caption{Summary of the region sampling and prompt construction (Sec. \ref{sampling_construct}) for each dataset.}
\label{tab:region_prompting}
\small
\begin{tabular}{p{2.3cm}p{2.8cm}p{2.8cm}p{3.2cm}p{4cm}}
\toprule
\textbf{Dataset} & \textbf{Region Source and Type} & \textbf{Selection Strategy} & \textbf{Margin Handling and Processing} & \textbf{Notes} \\
\midrule
VinDr-CXR~\cite{nguyen2022vindr} & Bounding boxes (thoracic abnormalities); 2D crops & Use all annotated bounding boxes with small context preservation & Expand bounding box by 10\%; resize crop to fixed input size while keeping aspect ratio & Focus on capturing localized opacities and consolidations within thoracic regions \\
\midrule
LIDC-IDRI~\cite{armato2011lidc} & 3D segmentation masks (lung nodules); 2D slices from mask & Select axial slice with highest radiologist agreement; extract tight bounding box around mask & No margin expansion; resize crop to fixed input size while keeping aspect ratio & Emphasizes fine-grained pulmonary nodule features such as spiculation, density variations, and margin sharpness \\
\midrule
SkinCon~\cite{daneshjou2022skincon} & Bounding boxes (cutaneous lesions); 2D crops from photographs & Use all visible lesion annotations; preserve color texture cues from skin & Expand bounding box by 15\% for context; resize crop to fixed size while preserving RGB fidelity & Prioritizes dermatological features such as lesion shape, color, boundary regularity, and skin texture context \\
\bottomrule
\end{tabular}
\end{table*}

To maintain evaluation diversity, we sample up to five regions per image where applicable. For images containing multiple abnormalities, 
regions are randomly selected to balance anatomical locations and pathology types. Regions are resized to a standard input size while preserving aspect ratio, ensuring consistent processing across all the datasets.

\subsection{Attribute Question Construction}
\label{attribute_question}

Medical image datasets typically lack region-level captions, making traditional text-based evaluation metrics infeasible for our task. Instead, we adopt an attribute-level verification strategy inspired by DLC-Bench~\cite{lian2025anythingdetailedlocalizedimage}, circumventing the need for ground-truth descriptions as references. For each sampled region, we design a set of clinically relevant binary (yes/no) questions that test whether the generated description correctly reflects the specific attributes of the localized abnormality. 

The questions are constructed based on the available annotations and the semantic characteristics of each dataset. Positive questions verify whether expected findings are correctly described, while negative questions ensure that unrelated or hallucinated attributes are not mentioned.

For VinDr-CXR, questions focus on the presence or absence of radiological features such as consolidation, lung opacity, or pleural effusion within the annotated bounding box. For LIDC-IDRI, questions center around pulmonary nodule properties, including margin sharpness (smooth vs. spiculated), internal density (solid vs. non-solid), and size-related descriptors. For SkinCon, since the original dataset lacks region-level annotations, we generate bounding boxes ourselves using a lightweight lesion detection method~\cite{wang2024achieving} as part of our flexible region-of-interest (ROI) pipeline. Based on these detected regions, we construct verification questions focused on key dermatological attributes such as lesion shape, color, border regularity, and surface texture, enabling clinically meaningful evaluation of region-level descriptions. These attributes are commonly used in clinical dermatology for diagnosis, thus serving as reliable semantic anchors for factual verification. Table~\ref{tab:attribute_questions} summarizes the attribute categories and corresponding evaluation focus for each dataset.

\begin{table*}[t]
\centering
\caption{Summary of the attribute verification tasks (Sec. \ref{attribute_question}).}
\label{tab:attribute_questions}
\small
\resizebox{\textwidth}{!}{
\begin{tabular}{p{2.5cm}p{3.2cm}p{4.2cm}p{4.2cm}}
\toprule
\textbf{Dataset} & \textbf{Attribute Type} & \textbf{Positive QA Example} & \textbf{Negative QA Example} \\
\midrule
VinDr-CXR~\cite{nguyen2022vindr} & Radiological findings (opacity, consolidation, effusion) & Is there increased opacity in the lower lobe of the lungs? & Is pneumothorax incorrectly mentioned in the localized description? \\
\midrule
LIDC-IDRI~\cite{armato2011lidc} & Pulmonary nodule morphology (margin, density, size) & Does the description mention a spiculated nodule margin? & Is lobulation falsely described when it is not present? \\
\midrule
SkinCon~\cite{daneshjou2022skincon} & Dermatological lesion characteristics (color, shape, texture) & Does the caption describe an irregular border and reddish hue? & Is scaling or ulceration incorrectly attributed to the lesion? \\
\bottomrule
\end{tabular}
}
\end{table*}

\subsection{Baselines and Evaluation Metrics}

\begin{figure*}[t]
    \centering
    \includegraphics[width=\textwidth]{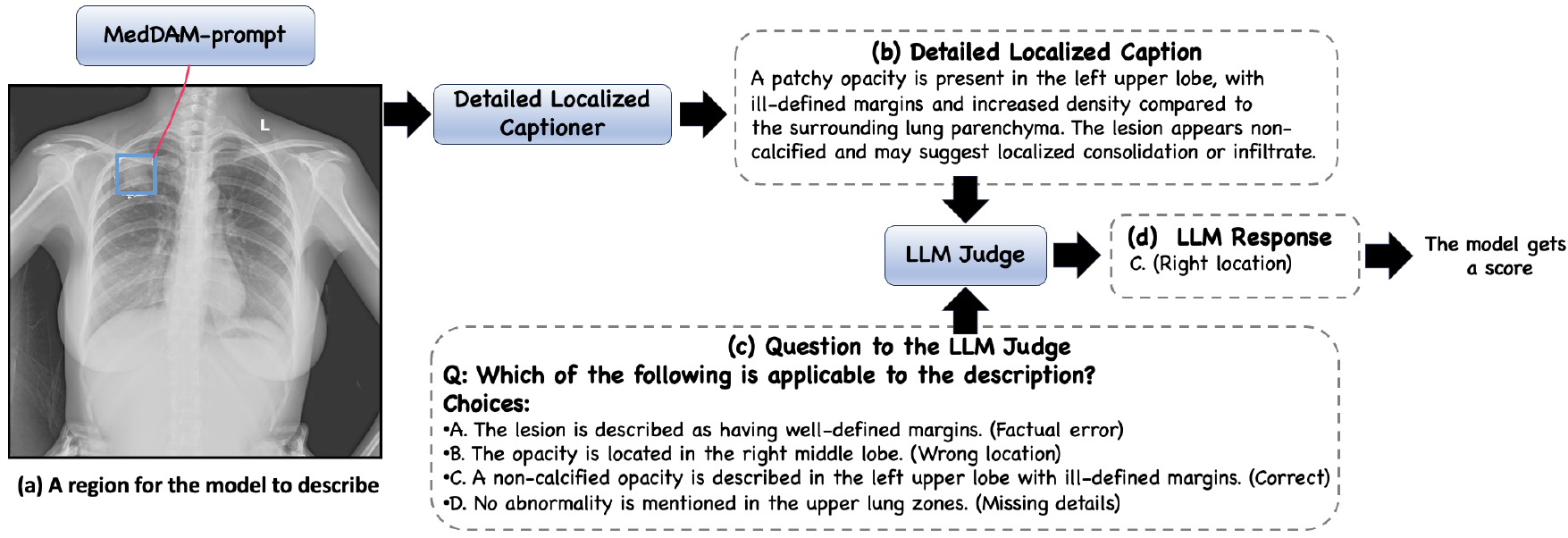}
    \vspace{-1em}
    \caption{
    \textbf{An example evaluation pipeline (i.e., calculating MedDLC-score).}
    (a) A region of interest is marked in a chest X-ray image and then used as input to MedDAM, prompted by a task-specific instruction.
    (b) The model generates a region-specific caption describing the abnormality within the marked region.
    (c) A question-answering task is set up to verify the factual accuracy and localization consistency of the generated caption, based on domain-specific attributes.
    (d) An LLM-based evaluator assigns a correctness label to the answer, and the model receives a score if the description matches the ground-truth semantic attributes.
    This framework enables reference-free benchmarking of fine-grained regional captioning performance on medical image understanding. \textbf{Although both the widely used LLM-score and our MedDLC-score involve a LLM Judge, the former is more effective for natural vision-language scenarios while the latter is tailored to the proposed task in medical domain.}
    }
    \label{fig:medical_dlc_pipeline}
\end{figure*}

We compare MedDAM against a series of state-of-the-art adaptable large VLMs that can understand visual content via texts. 
Specifically, we evaluate GPT-4o, Claude 3.7 Sonnet, LLaMA-3.2 Vision, Qwen2.5-VL, GPT-4Rol and OMG-LLaVA. These models represent the current frontier of vision-language pretraining and instruction tuning, covering a diverse range of architectures and training paradigms. For all the baselines, we provide region-level crops as inputs and retain access to the full-scale image input, as we do for MedDAM, prompting the models to generate region-specific descriptions. An example evaluation pipeline is illustrated in Fig. \ref{fig:medical_dlc_pipeline}.

To measure model performance, we adopt two protocols. First, we use the LLM-score, in which an independent and strong LLM (i.e., GPT-4o) serves as a judge to assess the fluency, relevance, factual correctness, and clinical plausibility of the generated descriptions. Each factor is rated individually, and the final score is computed by averaging across all evaluated regions. Second, we propose a clinically grounded and reference-free evaluation protocol, namely MedDLC-score,  tailored to medical semantic attributes. Instead of relying on ground-truth captions, it formulates a set of attribute-level binary verification tasks, assessing whether the generated descriptions accurately capture key clinical features such as lesion location, morphological appearance, and radiological findings. Model performance is reported as accuracy over positive and negative verification questions. This dual evaluation strategy facilitates a comprehensive assessment of both linguistic quality and medical factuality, yielding valuable feedback on the capabilities and limitations of region-specific captioning models in medical image understanding.


\subsection{Region-specific Prompting}
\label{prompting_scheme}

\begin{figure}[h!]
    \centering
    \includegraphics[width=\linewidth]{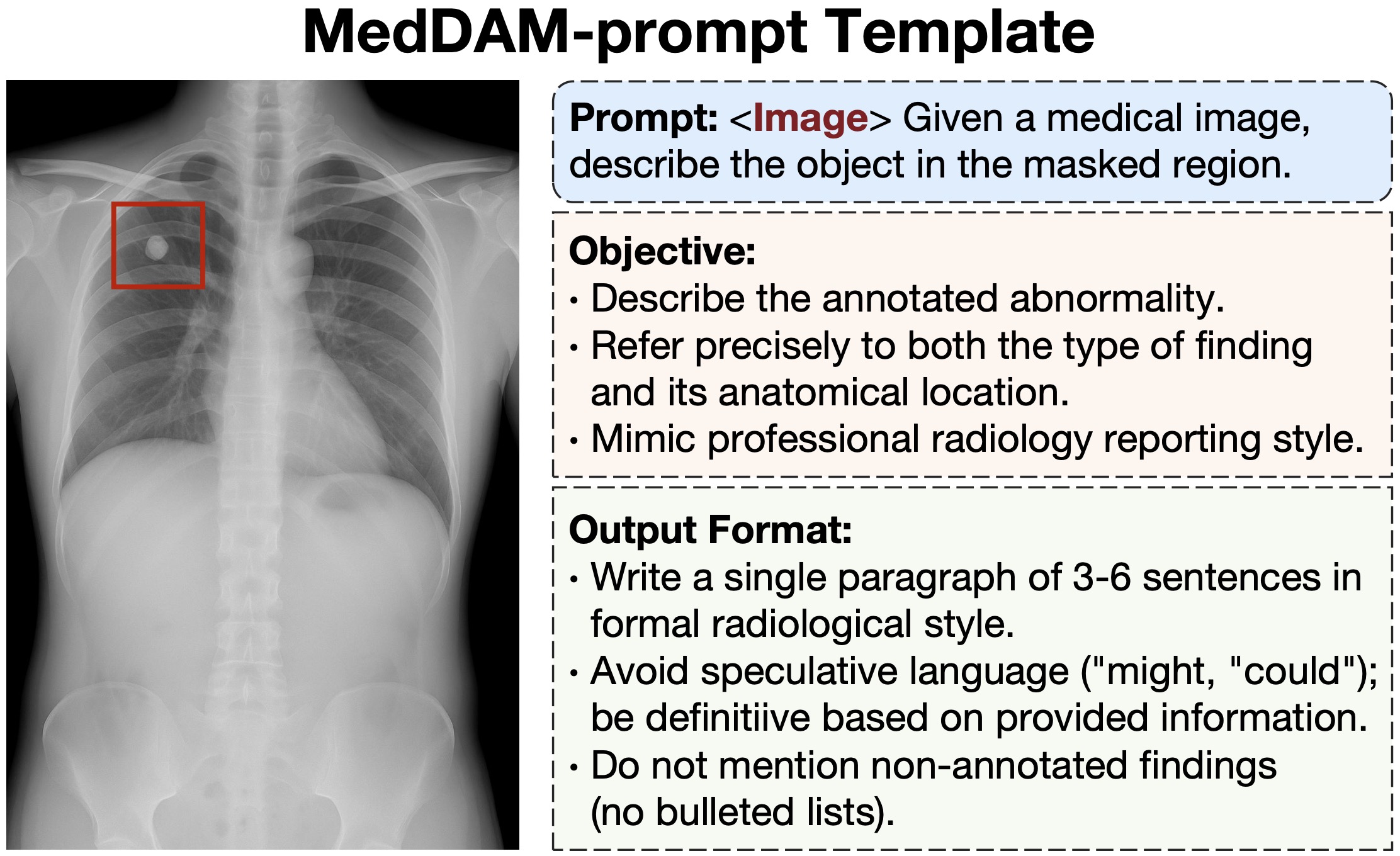}
    \vspace{-2em}
    \caption{
    \textbf{MedDAM-prompt Template.} This prompt guides the model to produce region-specific, clinically accurate descriptions by incorporating task constraints such as anatomical focus, output format, and information grounding. It is essential for adapting general captioning models like DAM to medical images.
    }
    \label{fig:medicaldam_prompt}
\end{figure}

A key component of realizing MedDAM is the task-specific prompting scheme, namely MedDAM-prompt, tailored to obtaining region-specific captions leveraging pretrained large VLMs. Unlike general captioning prompts, MedDAM-prompt explicitly instructs the model to (i) focus only on the annotated region, (ii) use anatomically precise terminology, and (iii) follow professional clinical report style while avoiding speculative or irrelevant content.
As shown in Fig.~\ref{fig:medicaldam_prompt}, the prompt includes explicit instructions that define the output format and customized objective. It is designed to minimize hallucinations and ensure the output's high relevance to the localized region. 

\subsection{Main Results}

\begin{table*}[t]
\centering
\caption{Main results on LLM-score and MedDLC-score. All models are evaluated in a zero-shot fashion. \textbf{Notably, following the practice in \cite{lian2025anythingdetailedlocalizedimage}, each score obtained from a model is the average result across the three datasets used in the experiment.}}
\label{tab:main_results}
\resizebox{\textwidth}{!}{
\begin{tabular}{lccccc}
\toprule
\textbf{Model} & \textbf{Type} & \textbf{LLM-score (↑)} & \textbf{MedDLC-score (↑)} & \textbf{Pos QA (↑)} & \textbf{Neg QA (↑)} \\
\midrule
GPT-4o~\cite{openai2024gpt4o}              & General       & \textbf{81.5} & 50.2 & 52.0 & 48.4 \\
Claude 3.7 Sonnet~\cite{anthropic2025claude} & General       & 79.2 & 47.5 & 49.0 & 46.0 \\
LLaMA-3.2 Vision~\cite{metallama}       
& General       & 75.3 & 43.8 & 45.1 & 42.5 \\
Qwen2.5-VL~\cite{bai2025qwen2}             & General       & 73.4 & 41.9 & 43.2 & 40.6 \\
GPT-4Rol~\cite{zhang2023instruction}       & Region-specific       & 77.1 & 45.7 & 47.3 & 44.0 \\
OMG-LLaVA~\cite{zhang2024omg}              & Region-specific  & 76.5 & 46.1 & 47.8 & 44.5 \\
\midrule
\textbf{MedDAM (Ours)}                 & Region-specific  & 78.9 & \textbf{63.6} & \textbf{65.1} & \textbf{62.0} \\
\bottomrule
\end{tabular}
}
\end{table*}

In Table~\ref{tab:main_results}, we  report the performance of general-purpose (but adaptable) and region-specific large VLMs on the proposed task. Evaluation metrics include the LLM-score, which assesses overall language quality, and our MedDLC-score, which reflects clinically grounded, region-specific captioning performance. The latter is further decomposed into positive and negative verification accuracy to provide a more granular analysis. 

Our MedDAM achieves the highest score on MedDLC-score (i.e., 63.6\%), significantly outperforming general models like GPT-4o (i.e., 50.2\%) and Claude 3.7 Sonnet (i.e., 47.5\%). These results underscore the effectiveness of the region-specific prompting scheme (Sec. \ref{prompting_scheme}) and the benefit of adapting general localized captioning to medical images. Notably, MedDAM excels in both positive (i.e., 65.1\%) and negative (i.e., 62.0\%) question types, indicating its ability to not only identify salient findings but also avoid hallucinating ungrounded content—a frequent failure case in general large VLMs.

Although MedDAM shows a comprehensive advantage over the baseline models, it trails GPT-4o in terms of LLM-score (i.e., 78.9\% vs. 81.5\%). It is not surprising since this score is calculated based on a powerful LLM as a judge, and in this experiment we leverage GPT-4o itself as the judge to evaluate its own performance, inevitably incurring bias. In addition to demonstrating the superiority, the results also reveal that MedDAM is capable of generating clinically meaningful and precise descriptions for localized regions, partially facilitated by the proposed evaluation protocol (i.e., MedDLC-score), which uniquely involves clinical significance in evaluating region-specific captioning for medical images. Moreover, MedDAM is free of any ground-truth region-text information as reference. Such a property is highly desired in medical domain, where data annotation could be extremely expensive. 







\section{Conclusion and Future Work}

In this work, we present \textbf{MedDAM}, the first framework leveraging the most recent breakthrough, namely Describe Anything Model (DAM), for region-specific captioning in medical images. We show that key to the realization MedDAM includes a proper design of text prompts by medical experts, and an evaluation benchmark to logically assess MedDAM and its competitors. To establish this benchmark, we develop an evaluation protocol tailored to medical images, a data pre-processing pipeline to capture region-level cues from images, and a QA template library, jointly evaluate the performance without resorting to ground-truth descriptions that are scarce in medical images. 
Our experiments on three publicly available datasets—VinDr-CXR, LIDC-IDRI, and SkinCon—demonstrate that MedDAM significantly improves factual alignment and regional specificity over strong and adaptable large VLMs in zero-shot fashion, also revealing that there is much leeway for boosting the performance of  modern VLMs on medical image understanding. 

Looking ahead, we envision several promising directions. First, we plan to \textbf{fine-tune DAM on large-scale, weakly annotated medical datasets} using pseudo-labeling or self-supervised alignment schemes to further improve the accuracy of generated region-specific captions. 
Second, we plan to extend MedDAM to additional modalities and specialties, including ophthalmology and pathology,  enabling broader benchmarking across various medical image understanding tasks. Finally, it would be interesting to integrate structured knowledge (e.g., RadLex~\cite{radlex}, SNOMED~\cite{snomedct}) into the prompting and evaluation schemes to enhance interpretability and domain alignment of the MedDAM. We expect that this work will inspire more future research on region-specific medical image understanding to eventually benefit clinical applications.

\bibliography{example_paper}
\bibliographystyle{icml2025}



\end{document}